\documentclass[sigconf]{acmart}

\usepackage{algorithm}
\usepackage{algpseudocode}
\usepackage[normalem]{ulem}
\usepackage{cleveref}
\crefname{algorithm}{Algorithm}{Algorithms}
\Crefname{algorithm}{Algorithm}{Algorithms}

\AtBeginDocument{%
  }

\copyrightyear{2026}
\acmYear{2026}
\setcopyright{cc}
\setcctype{by}
\acmConference[GECCO '26]{The Genetic and Evolutionary Computation Conference}{July 13--17, 2026}{San José, Costa Rica}
\acmBooktitle{The Genetic and Evolutionary Computation Conference (GECCO '26), July 13--17, 2026, San José, Costa Rica}
\acmDOI{10.1145/3795095.3805157}
\acmISBN{979-8-4007-2487-9/2026/07}

\begin{document}

\title{A Comparative Study of Model Selection Criteria for Symbolic Regression}


\author{Ali Soltani}
\affiliation{%
  \institution{Aarhus University}
  \department{Department of Mechanical Engineering}
  \city{Aarhus}
  \country{Denmark}}
\email{asoltani@mpe.au.dk}
\orcid{0009-0007-8588-7294}

\author{Gabriel Kronberger}
\affiliation{%
  \institution{Heuristic and Evolutionary Algorithms Laboratory (HEAL)}
  \department{University of Applied Sciences Upper Austria}
  \city{Hagenberg}
  \country{Austria}}
\email{gabriel.kronberger@fh-hagenberg.at}

\author{Fabricio Olivetti de França}
\affiliation{%
  \institution{Federal University of ABC}
  \department{Center for Mathematics, Computing, and Cognition}
  \country{Brazil}}
\email{folivetti@ufabc.edu.br}

\author{Mattia Billa}
\affiliation{%
  \institution{University of Modena and Reggio Emilia}
  \department{Department of Physics, Informatics and Mathematics}
  \city{Modena}
  \country{Italy}}
\email{mattia.billa@unimore.it}
\orcid{0009-0005-1979-8918}

\author{Alessandro Lucantonio}
\affiliation{%
  \institution{Aarhus University}
  \department{Department of Mechanical Engineering}
  \city{Aarhus}
  \country{Denmark}}
\email{a.lucantonio@mpe.au.dk}
\orcid{0000-0002-9807-5451}

\renewcommand{\shortauthors}{Soltani et al.}

\begin{abstract}
  Effective model selection is critical in symbolic regression (SR) to identify mathematical expressions that balance accuracy and complexity, and have low expected error on unseen data. Many modern implementations of genetic programming (GP) for SR generate a set of Pareto optimal candidate solutions, but reliable automatic selection of solutions that generalize well remains an open issue. Current literature offers various information-theoretic and Bayesian approaches, yet comprehensive comparisons of their performance across different data regimes are limited.  This study presents a systematic empirical comparison of widely used selection criteria: the Akaike information criterion (AIC), the corrected AIC (AICc), the Bayesian information criterion (BIC), minimum description length (MDL), as well as Efron's bootstrap estimate for the in-sample prediction error on seven synthetic datasets with Gaussian noise. We rank candidate expressions generated by perturbing ground-truth functions to assess generalization error and selection probability of the ground-truth expression. Our findings reveal that MDL consistently identifies models with the lowest test error and the shortest length across most datasets. While no single criterion dominates all results, MDL and BIC produced the highest probability of selecting the ground-truth expressions. 

\end{abstract}




\maketitle


\section{Introduction}
Symbolic regression (SR) is a form of equation discovery from data, and
Genetic programming (GP) has emerged as a successful approach to solve this task~\cite{Koza1992, Langdon2002}. GP algorithms typically generate a population of candidate solutions, represented as tree-like structures encoding mathematical expressions. Over successive generations, the population evolves toward better-performing expressions, leveraging principles inspired by natural selection. GP's inherent flexibility allows for multi-objective optimization, where solutions can effectively balance between accuracy and complexity. This balance is especially crucial in practical applications, where model interpretability and predictive performance are equally important. This way, GP produces a Pareto front, multiple optimal solutions that tradeoff between these competing objectives, providing valuable insights for selecting interpretable models that generalize well.
Although symbolic regression algorithms try to return an optimized model in terms of complexity and accuracy, selecting the best model in terms of generalization is still challenging. 
Therefore, effective model selection criteria are essential \cite{nicolau2021choosing, maciel2016measuringcomplexity}.

Several established model selection methods have been proposed, including information-theoretic criteria, such as the Akaike information criterion (AIC) and the Bayesian information criterion (BIC). The minimum description length (MDL) principle provides another robust framework that balances model complexity against fit by encoding both the model and data. Additionally, cross-validation approaches have been widely used for estimating the generalization performance by partitioning data into training and validation subsets \cite{arlot2010}. 

Comprehensive comparisons of selection criteria within SR remain limited. One of the few examples is \cite{Montana2011}, in which AIC, BIC and structural risk minimization were compared on a set of synthetic problems. Another recent example is~\cite{Ramlan2026}. 
However, practical guidance on which criterion to prefer under which conditions is still incomplete.
In this work, we evaluate and compare model selection criteria for SR, including the AIC, the corrected Akaike information criterion (AICc), BIC, MDL, and Efron's bootstrap-based covariance-penalty estimate of the in-sample prediction error ($\rm Err_{in}$). We evaluate the performance of these criteria on synthetic benchmarks consisting of seven noisy datasets generated from functions commonly used in the GP literature by comparing their ranking ability with
respect to the ranking of the models based on the test error. More specifically, the criteria studied here are intended to rank and select among candidate expressions produced at the end of an SR algorithm run, without being tied to any specific SR algorithm. Overall, our results offer practical guidance on choosing SR models that generalize well, and shed light on the strengths and weaknesses of different selection approaches.

The paper is structured as follows: We first describe the methodology, including a description of the model selection criteria used, the used datasets, and our evaluation procedure. After the presentation of the empirical results and a discussion of findings and the limitations of our work, we specifically connect to related work and discuss how our findings align with prior work. Finally, we present our conclusions and the wider implications.

\section{Methods}
\label{sec:headings}

\subsection{Datasets}
We create training and test datasets for each of the functions in Table ~\ref{tab:benchmark-functions} to estimate the generalization error of the candidate expressions and assess the performance of the model selection criteria in identifying models that generalize well. 
The benchmark functions  in Table ~\ref{tab:benchmark-functions} were selected because they exhibit different characteristics including nonlinearity, non-separability, dimensionality, and structural complexity.
The training datasets consist of 100 randomly distributed points within each of the intervals reported in Table ~\ref{tab:benchmark-functions}. The corresponding targets are obtained by perturbing the ground-truth targets ($y$) with Gaussian noise:
\begin{equation}
    y_{\text{noisy}} = y + \varepsilon,\quad \varepsilon \sim \mathcal{N}(0, (0.1\,\sigma(y))^2\text{I}).
    \label{eq:noise-level}
\end{equation}
The test data points (10000) are sampled within the same intervals as for the training data points, but the test targets are noise-free.

\subsection{Candidate expressions}
To evaluate the model selection criteria independently of any specific SR algorithm, a controlled pool of candidate expressions is generated by perturbing the ground-truth functions in Table \ref{tab:benchmark-functions}. 
To create a robust set of alternatives for the selection process, a mutation procedure selects a node uniformly at random from the ground-truth tree and replaces it with a random subtree of depth between 2 and 10. These subtrees are composed of binary operators ($+$, $-$, $*$, $/$, $\rm pow$), unary operators ($\sin$, $\cos$, $\sqrt{\cdot}$, $\exp$), features, and numerical parameters, collected in the array $\theta$. To handle the overflow and division by zero, a protected version of ($/$, $\rm pow$, $\sqrt{\cdot}$, $\exp$) is used. The protected exponential and power are capped versions of the corresponding standard operators, the protected square root returns $\sqrt{|\cdot|}$, and protected division is defined as follows, with $\tau = 10^{-6}$:
\begin{equation}
\operatorname{protected-div}(l,r)=
\begin{cases}
\dfrac{l}{r}, & |r| > \tau,\\
l, & \text{otherwise}.
\end{cases}
\end{equation}
For each mutated structure, the associated free numerical parameters are fitted to training data using the BFGS optimization algorithm. To avoid local optima, the optimization process is repeated 100 times with different starting points, and the parameter values corresponding to the lowest mean squared error on the training set ($\mathrm{MSE}_{\text{train}}$) are selected. This mutation and fitting process is repeated until it produces 100 perturbed trees with a $\mathrm{MSE}_{\text{train}}$ lower than the original ground-truth model, thus generating overfitted models. 
This procedure allows for an assessment of whether selection criteria can successfully retrieve the ground-truth model from a set of overfitted alternatives. 

\begin{table*}
    \centering
    \renewcommand{\arraystretch}{1.5}
    \begin{tabular}{cp{2cm}ll}
      Id & Ref. & Expression & Distributions \\
      \hline
      $f_1$ & \cite{Friedman1983,Kronberger2024} & y = $10\, \sin(\pi\, x_1\, x_2) + 20\, (x_3 - 0.5)^2 + 10\, x_4 + 5\, x_5$ & $x \sim U(0, 1)^{10}$\\
      $f_2$ & \cite{Vladislavleva2009,Smits2005} & y = $e^{-(x_1 - 1)^2}\, \left(1.2 + (x_2 - 2.5)^2\right)^{-1}$ & $x \sim U(0, 4)^2$ \\ 
      $f_3$ & \cite{Vladislavleva2009,Salustowicz1997} & y = $e^{-x}\, x^3 \cos(x)\, \sin(x)\, \left(\cos(x)\, \sin(x)^2 - 1\right)$ & $ x \sim U(0, 10) $ \\ 
      $f_4$ & \cite{Vladislavleva2009} & y = $ f_3(x_1)\, (x_2 - 5)$ & $x \sim U(0, 10)^2$ \\ 
      $f_5$ & \cite{Vladislavleva2009,Topchy2001} & y = $30\, (x_1 - 1)\, (x_3 - 1) (x_1\, x_2^2 - 10\, x_2^2)^{-1}$ & y = $x_{1,3}\sim U(0.05,2)^2$ \\ 
            & &                                                              & $x_2\sim U(1,2)$ \\ 
      $f_6$ & \cite{Vladislavleva2009} & y = $\frac{(x_1 -3)^4 + (x_2 - 3)^3 - (x_2 -3)}{(x_2 - 2)^4 + 10 }$ & $x \sim U(0, 6)^2$ \\ 
      $f_7$ & \cite{Vladislavleva2009,Topchy2001} & y = $(x_1-3)\, (x_2-3) + 2 \sin\left( (x_1 - 4)\, (x_2 - 4)\right)$ & $ x \sim U(0,6)^2$ \\ 
      \hline
    \end{tabular}
    \caption{Synthetic benchmark functions and sampling distributions used to evaluate the model selection criteria.}\label{tab:benchmark-functions}
    
\end{table*}

\subsection{Model selection criteria}
\label{sec:criteria}
In the following, we report the definitions of the model selection criteria that we compare in this work.

\subsubsection{Akaike information criterion}
This criterion trades off model accuracy and complexity via the following equation
\begin{equation}
    \mathrm{AIC} = -2 \log \mathcal{L}(\hat{\theta}) + 2p,
    \label{eq:AIC}
\end{equation}
where $\log \mathcal{L}$ is the log-likelihood, $\hat{\theta}$ is the vector of length $p$ representing the maximum likelihood estimate of the model parameters. 
By assuming errors to be independent and normally distributed with common variance $\hat{\sigma}^2$, the log-likelihood for $n$ observations may be computed as: 
\begin{equation}
    \log \mathcal{L} (\theta)
    \;=\;
    -\frac{n}{2}\,\log (2\pi\hat{\sigma}^{2})
    \;-\;
    \frac{1}{2\hat{\sigma}^{2}}
    \sum_{i=1}^{n}
    [y_i - \mu(x_i;\theta)]^{2},
    \label{eq:LL}
\end{equation}
where $y_i$ are the labels, $x_i$ are the samples, and $\mu$ is the model. For model selection, we evaluate ~\eqref{eq:AIC} for each model on the training set, with $\hat{\sigma}=0.1\sigma(y)$ in \eqref{eq:LL} 
and select the models with smallest AIC values. 

\subsubsection{Corrected Akaike information criterion}
The AIC is a first-order bias correction and therefore tends to select models that are too complex whenever the effective sample size is modest relative to the number of estimated parameters. \citet{BurnhamAndersonBook2004} proposed a second-order correction as
\begin{equation}
    \mathrm{AIC}_c = \mathrm{AIC} + \frac{2p(p+1)}{n - p - 1},
    \label{eq:aicc}
\end{equation}
where the extra term becomes negligible as $n/p \rightarrow \infty$, but is essential for smaller datasets, say when $n/p < 40$, to reduce the optimism of AIC. 

\subsubsection{Bayesian information criterion} Introduced by Schwarz \cite{schwarz(BIC)}, it is derived from a large-sample approximation to the Bayesian model evidence (marginal likelihood). The BIC is defined as:
\begin{equation}\mathrm{BIC} = -2 \log \mathcal{L}(\hat{\theta}) + p \log n.
\label{eq:BIC}
\end{equation}
Similarly to the AIC, the BIC balances goodness-of-fit with model parsimony. However, the BIC replaces the constant penalty of $2p$ with a penalty that grows logarithmically with the sample size. For $n > e^2$, where $e$ is Euler’s number, the BIC penalizes additional parameters more severely than the AIC, typically resulting in more concise, interpretable formulas.

\subsubsection{Minimum description length}
The MDL principle proposes that, among competing models, we should prefer the one that yields the shortest total code length, that is, the most concise joint encoding of the model and the data it explains \citep{grunwald(MDL), RISSANEN1978(MDL)}. Formally, the total code length is defined as
\begin{equation}
    L(D)=L(M)+L(D\mid M),
    \label{eq:mdl-basic}
\end{equation}
where $L(M)$ denotes the number of bits required to transmit the model itself and $L(D\mid M)$ is the number of bits needed to encode the data once the receiver knows~$M$. 
In SR, $L(M)$ must account for both the structure of the expression and its numeric parameters.  A practical formula, suggested by \citet{bartlett2023exhaustive(MDL)}, reads
\begin{equation}
\begin{split}
    L(D) \;=\;&
    -\log \mathcal{L}(\hat{\theta})
    \;+\; \omega\log u
    \;-\;\frac{p}{2}\log 3
    \;+\;\sum_{j} \log c_{j} \\
    &+\;\sum_{i=1}^{p} \Bigl[\tfrac12\log I_{ii}
    + \log |\hat{\theta}_{i}|\Bigr],
\end{split}
\label{eq:mdl-bartlett}
\end{equation}
with the following definitions:
\begin{itemize}
    \item[$\omega$] Number of nodes in the expression tree;   
    \item[$u$] Number of distinct symbols (\textit{i.e.} operators, variables, and parameters) appearing in the expression tree; 
    \item[$c_{j}$] Constants occurring in the model;
    \item[$I_{ii}$] $i$-th diagonal element of the observed Fisher-information matrix computed at $\hat{\theta}$.
\end{itemize}
Here the code length for the parameter estimates relates to an improper Bayesian prior \cite{Bartlett_2023}.
Unlike criteria such as AIC or BIC that penalize complexity only through the count of parameters this formulation also accounts for the structural complexity and the precision of parameter estimates. MDL measures complexity in the same information units (nats) as data misfit, eliminating the need for an arbitrary weighting hyperparameter.

\subsubsection{In-sample error}
The training error is an optimistic measure of the generalization error. Part of this optimism can be explained by considering the quantity \cite{Efron2004(Err_in)}
\begin{equation}
    \mathrm{Err}_i = \mathbb{E}_0(y_i^0-\hat{\mu}_i)^2,
\end{equation}
where the expectation is taken with respect to the training observations $y_i^0$ keeping the model predictions $\hat{\mu}_i = \mu(x_i;\,\hat{\theta})$ corresponding to the training samples $x_i$ fixed. The average of this quantity over the number of samples is the \textit{in-sample error} $\mathrm{Err}_{\rm in}$ \cite{hastie2009elements(Err_in)}. In particular, it can be shown that
\begin{equation}
\mathbb{E}(\mathrm{Err}_{\rm in}) = \mathbb{E}\left( \mathrm{MSE}_{\textrm{train}} + \frac{2}{n}\sum_{i=1}^n\mathrm{cov}\,(y_i, \hat{\mu}_i)\right), \label{eq:covpen}
\end{equation}
where the expectation is with respect to the data-generating distribution for the observed $y_i$, which implies that the model is refitted to different training sets. From this formula it is evident that we must add a \textit{covariance penalty} to the training error $\mathrm{MSE}_{\textrm{train}}$, in order to unbiasedly estimate $\mathrm{Err}_{\textrm{in}}$. 

The \textit{parametric bootstrap} method offers a way to estimate the covariance penalty term in \eqref{eq:covpen} as detailed in Algorithm ~\ref{alg:bootstrap_cov_i_bigmodel} \cite{Efron2004(Err_in)}. Notice that, differently from the other criteria, an estimate $\tilde\sigma$ of the noise variance was used, instead of the value $\hat{\sigma}$ of the data-generating distribution. After obtaining $\widehat{\mathrm{Cov}}_i$ via Algorithm~\ref{alg:bootstrap_cov_i_bigmodel}, the quantity $\textrm{Err}_{\textrm{in}}$ can be estimated as follows:
\begin{equation}
\widehat{\mathrm{Err}}_{\textrm{in}} \;=\; \mathrm{MSE}_{\textrm{train}} \;+\; \frac{2}{n} \sum_{i=1}^n \widehat{\mathrm{Cov}}_i.
\label{eq:insample}
\end{equation}
Intuitively, if changes in model predictions correlate with the noise added to training labels when the latter are perturbed, the model is capable of fitting the noise, and is thus overfitting, resulting in a higher covariance penalty. Hence, the estimate of the in-sample error in \eqref{eq:insample} can be used to rank models and perform model selection.

\begin{algorithm}[t!]
\caption{Parametric bootstrap estimate of the covariance penalty.}
\label{alg:bootstrap_cov_i_bigmodel}
\begin{algorithmic}[1]
\Require Training data $\{(x_i,y_i)\}_{i=1}^n$, candidate model $\mu(x;\theta)$, bootstrap replicates $B=200$.
\State Fit the candidate model parameters on $(x_i,y_i)$: obtain $\hat{\theta}$ and compute $\hat\mu_i=\mu(x_i;\hat\theta)$.
\State Fit a Random Forest regressor $m^{\text{RF}}$ on $(x_i,y_i)$ and compute $\bar{y}^{*}_i=m^{\text{RF}}(x_i)$.
\State Estimate the noise variance: 
\Statex \[
\tilde\sigma^2 \;=\; \frac{1}{n}\sum_{i=1}^n \bigl(y_i - \bar{y}^{*}_i\bigr)^2 .
\]
\For{$b=1$ to $B$}
    \State Draw $\hat{\varepsilon}^{*b}_i \overset{\text{i.i.d.}}{\sim} \mathcal{N}(0,\tilde\sigma^2)$ for $i=1,\dots,n$.
    \State Form synthetic responses $y^{*b}_i \,=\, \bar{y}^{*}_i \,+\, \hat{\varepsilon}^{*b}_i$. 
    \State Refit model parameters  on $\{(x_i,y^{*b}_i)\}_{i=1}^n$: obtain $\hat{\theta}^b$ and compute $\hat\mu^{*b}_i=\mu(x_i;\hat{\theta}^b)$. 
\EndFor
\State Compute $\bar{\mu}^{*}_i \,=\, \frac{1}{B}\sum_{b=1}^B \hat\mu^{*b}_i$.
\State Compute the covariance estimates:
\Statex \[
\widehat{\mathrm{Cov}}_i \;=\; \frac{1}{B-1}\sum_{b=1}^B \bigl(\hat\mu^{*b}_i - \bar{\mu}^{*}_i\bigr) \,\bigl(y^{*b}_i - \bar{y}^{*}_i\bigr) .
\]
\State \Return $\widehat{\mathrm{Cov}}_i$ for $i=1,\dots,n$.
\end{algorithmic}
\end{algorithm}

\subsection{Computational complexity analysis}
While the primary objective, as we will see in the following sections, is to identify models with high generalization performance, the computational overhead of these criteria is a significant factor in the search process of SR. This is especially relevant because of the iterative nature of GP, where thousands of candidate expressions may be evaluated per generation. The criteria compared in this study exhibit different scaling behaviors with respect to the number of samples $n$, the number of free parameters $p$, and the expression size $\omega$.

The information-theoretic criteria (but also $\mathrm{MSE}_{\text{train}}$), including the AIC, AICc, and BIC, are the most computationally efficient to compute once the maximum likelihood estimates $\hat{\theta}$ have been obtained. Their cost is dominated by the evaluation of the log-likelihood $\log \mathcal{L}(\hat{\theta})$, which requires a single pass over the training data, resulting in a complexity of $O(n)$. The subsequent penalty terms involve only scalar arithmetic on $p$ and $n$, adding a negligible $O(1)$ overhead.

MDL introduces additional complexity due to its requirement for the observed Fisher-information matrix. While the computation of the structural penalty terms scales with the number of nodes $\omega$, the dominant cost arises from computing the diagonal elements $I_{ii}$ of the Hessian of the log-likelihood. For an expression with $p$ parameters evaluated on $n$ samples, calculating these diagonal terms typically requires $O(n \cdot p)$ operations. While slightly more expensive than AIC and BIC, MDL remains highly efficient considering the usually small expressions in SR.

In contrast, the bootstrap-based in-sample error estimate ($\mathrm{Err}_{\rm in}$) is by far the most computationally intensive criterion. As detailed in Algorithm \ref{alg:bootstrap_cov_i_bigmodel}, the procedure requires fitting a ``big'' model (e.g., a Random Forest) to estimate the noise variance. Furthermore, the criterion requires $B$ bootstrap iterations, each involving a complete refitting of the parameters $\hat{\theta}^b$ of the candidate expression on the perturbed data. Given that parameter fitting in SR often uses iterative numerical optimization like BFGS, the cost for $\mathrm{Err}_{\rm in}$ scales as $O(B \cdot C_\text{fit} + \text{Cost}(m^\text{RF}))$, where $C_\text{fit}$ is the cost of the non-linear optimization process. With $B$ typically set to 100 or more to ensure stability, $\mathrm{Err}_{\rm in}$  can be several orders of magnitude slower than its counterparts, a factor that can limit its applicability in evolutionary search. 

\subsection{Evaluation procedure}
\label{sec:evaluation_metrics}
To assess the performance of the various model selection criteria, we employed four complementary metrics that evaluate generalization, ranking fidelity, and model parsimony:
\begin{itemize}
  \item \textbf{Average $\mathrm{MSE}_{\text{test}}$:}
  This metric computes the mean squared test error ($\text{MSE}_{\text{test}}$) of the top-$k$ models as ranked by a specific criterion. Lower values indicate that the criterion effectively prioritizes models with superior generalization performance, successfully identifying candidates that reside near the top of the true $\text{MSE}_{\text{test}}$ distribution.
  
  \item \textbf{Precision at $k$:} 
  Precision at $k$ measures the fraction of the selected top-$k$ models based on a given criterion that actually belong to the true top-$k$ 
  according to the $\mathrm{MSE}_{\text{test}}$. 
  A precision of 1.0 indicates a perfect overlap with the highest-generalizing models, whereas lower values reflect the inclusion of less accurate candidate expressions.
  \item \textbf{Average size:}
  This quantity evaluates the parsimony of the selected models by calculating the mean structural complexity (total number of nodes) of the top-$k$ expressions. This metric is essential for understanding how each criterion balances the trade-off between predictive accuracy and model interpretability.

  \item \textbf{Ground-truth hit:}
 This metric identifies the minimum rank $k$ at which the exact ground-truth generating expression is successfully recovered among the models prioritized by a given criterion. We report the ``first-hit" $k$ for each benchmark function to compare how efficiently different selection methods prioritize the correct model structure within the candidate pool.
  
\end{itemize}

\section{Results and discussion}


The empirical evaluation of the model selection criteria of Section~\ref{sec:criteria} across seven synthetic benchmarks reveals distinct trade-offs between predictive accuracy, model size, and ground-truth recovery. 
As a baseline for this evaluation, we consider the mean squared error on the training set ($\rm{MSE}_{\rm train}$), which provides a coarse and biased estimate of the test error $\rm{MSE}_{\rm test}$.
All curves show performance as a function of $k$ (the number of selected models by each criterion) on the same candidate pool per benchmark function. Our results demonstrate that  the choice of the model selection method fundamentally dictates the generalization capability of the selected expressions, while no single criterion dominates.

\paragraph{Average $\mathrm{MSE}_{\text{test}}$}

\begin{figure}[t]
    \centering
    \includegraphics[width=\columnwidth]{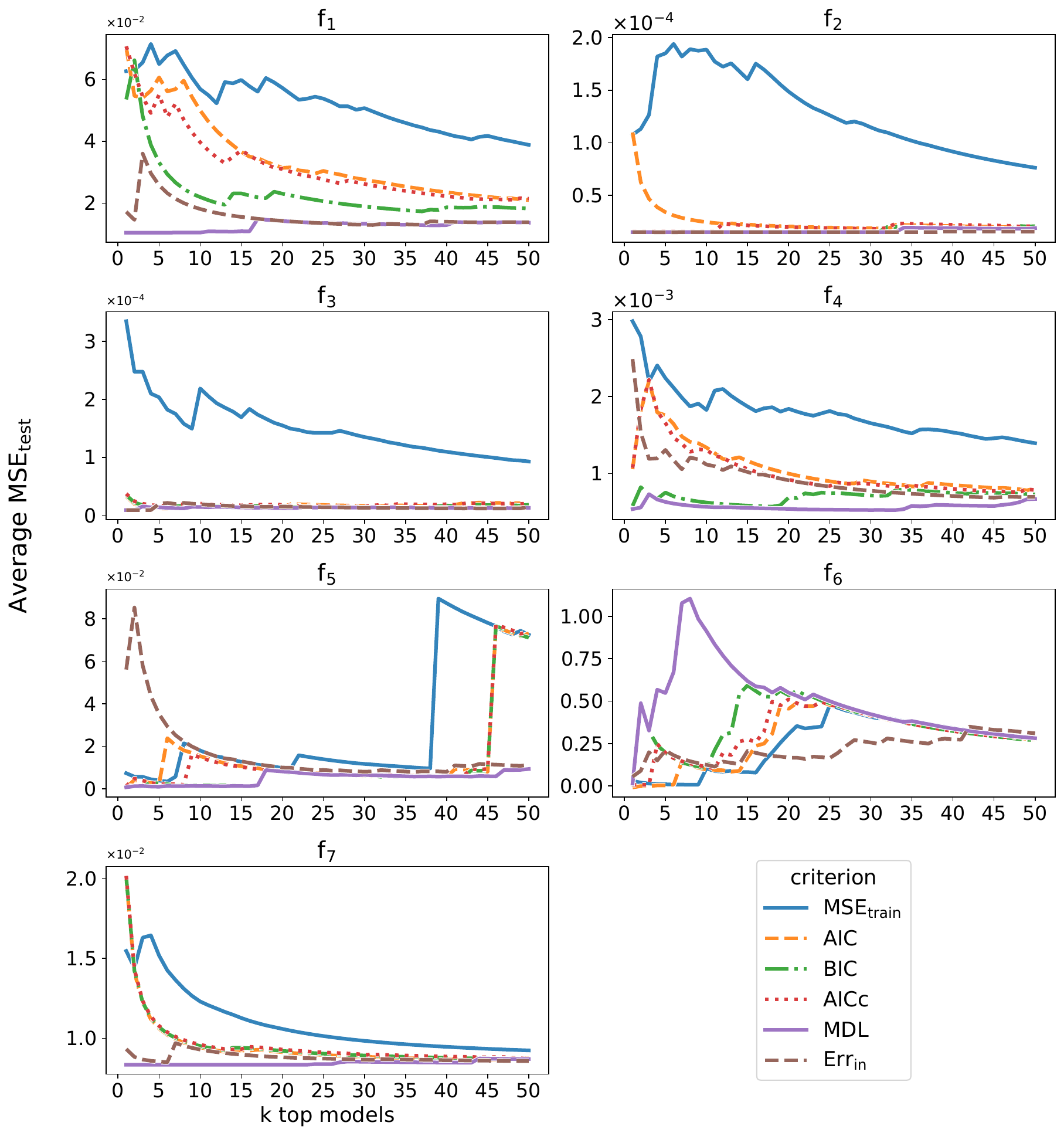}
    \caption{Average $\text{MSE}_{\text{test}}$ for different model selection criteria. The average test MSE is taken over the top-$k$ models, given on the $x$-axis.}
    \label{fig:grid_avg_test_err}
\end{figure}
 
The analysis of the average $\mathrm{MSE}_{\text{test}}$ (Figure \ref{fig:grid_avg_test_err}) highlights the limitations of using the unregularized training error for model selection. $\mathrm{MSE}_{\text{train}}$ consistently performs the worst across nearly all benchmarks. In contrast, MDL emerges as the most robust criterion, yielding the lowest or near-lowest test error for small $k$ in six of the seven benchmarks. This suggests that MDL’s joint encoding of model structure and parameters effectively suppresses the selection of overfitted candidates.
BIC serves as a stable intermediate choice, providing more consistent results across the benchmark suite than the AIC variants. While AICc shows a slight improvement over the standard AIC for $f_{1}$ and $f_{2}$, both generally rank lower than MDL and BIC in terms of generalization. The results related to $\textrm{Err}_{\textrm{in}}$ are mixed: while it achieves good performance for $f_{1}$, $f_{2}$, $f_{3}$, $f_{6}$, and $f_{7}$, it performs poorly on $f_{4}$ and $f_{5}$. We argue that the dependence of the $\textrm{Err}_{\textrm{in}}$ criterion on the RF model accuracy and the noise-variance estimate influences this behavior.

\paragraph{Average expressions size}

\begin{figure}[t]
    \centering
    \includegraphics[width=\columnwidth]{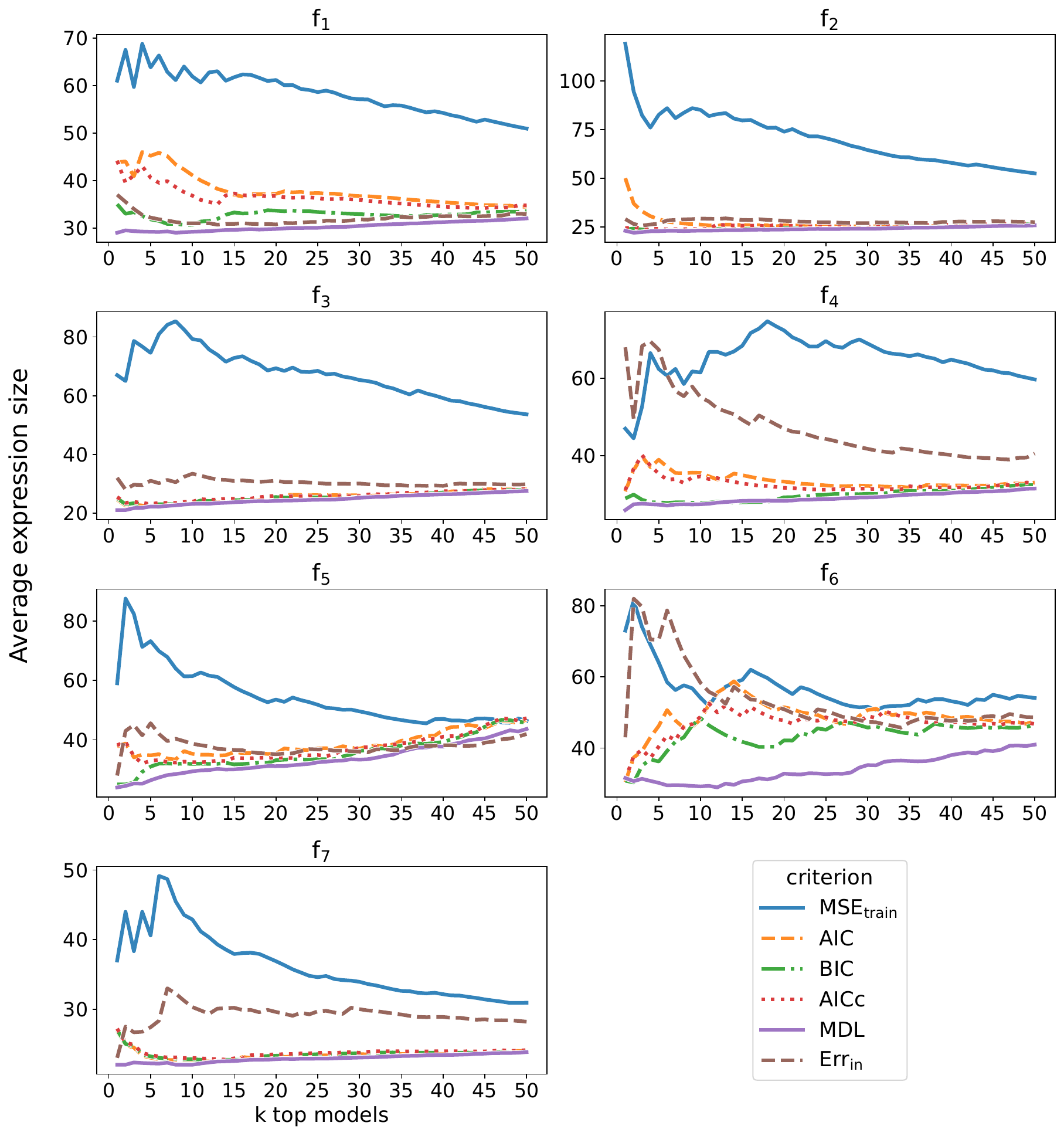}
    \caption{Average expression size for different model selection criteria.}
    \label{fig:grid_avg_size}
\end{figure}

Figure ~\ref{fig:grid_avg_size} shows the average expression size of the top-$k$ models chosen by each criterion. As expected, $\mathrm{MSE}_{\text{train}}$ selects the largest expressions. 
MDL and BIC consistently prioritize the most parsimonious expressions. MDL’s explicit structural penalty results in the smallest expressions in the top ranks, which, when coupled with its low average $\mathrm{MSE}_{\text{test}}$, indicates a superior accuracy-simplicity trade-off. AIC and AICc occupy a middle ground, though AICc's second-order correction results in slightly more compact models than standard AIC in some cases. $\text{Err}_\text{in}$ often yields larger expressions, particularly in $f_{4}$ and $f_{5}$, which contributes to its poor generalization on those datasets. However, for $f_{6}$, $\text{Err}_\text{in}$ selects larger models such that it can capture the underlying data shape more effectively than the more conservative MDL.

\paragraph{Precision at $k$}

\begin{figure}[t]
    \centering
    \includegraphics[width=\columnwidth]{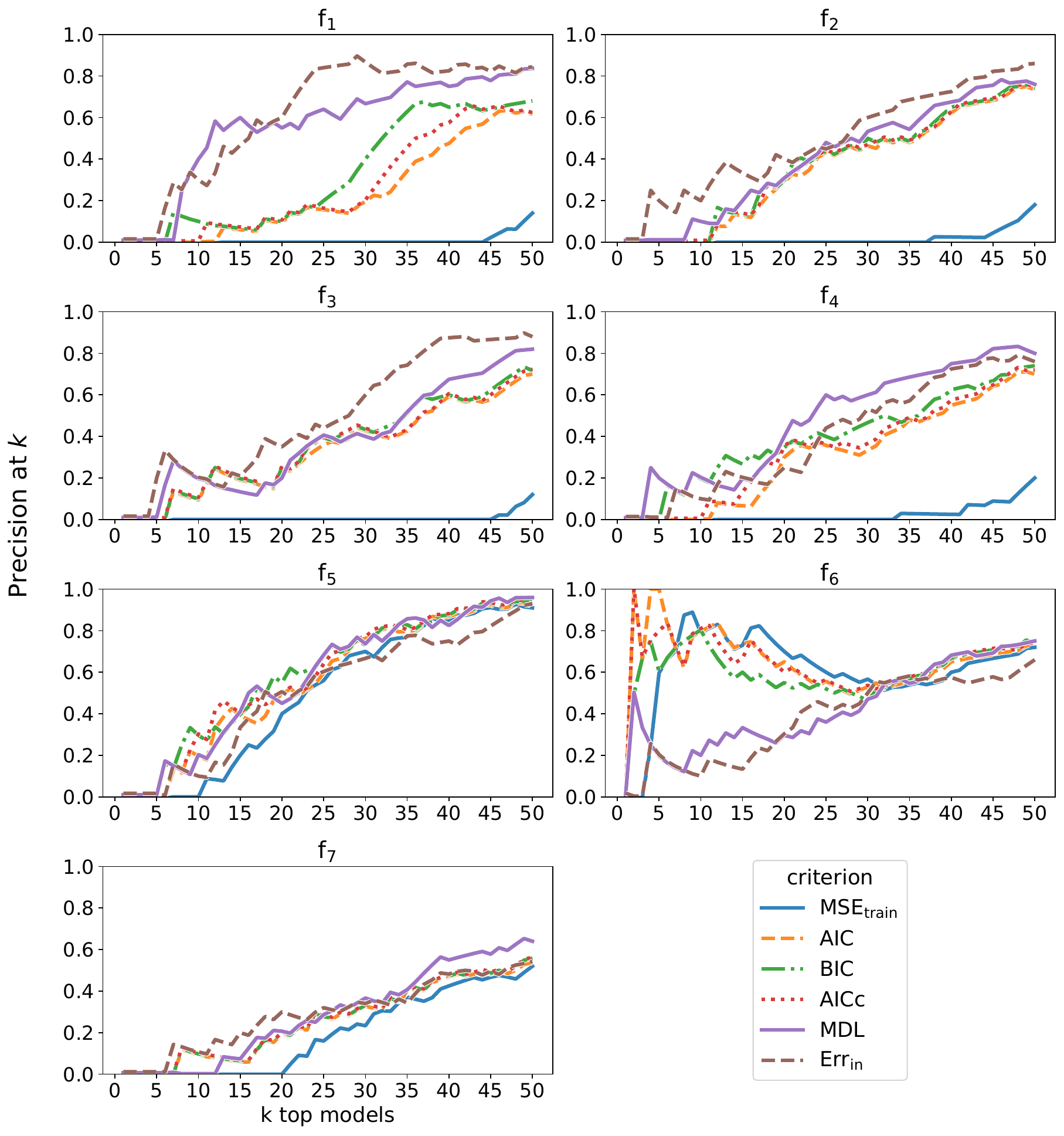}
    \caption{Precision at $k$ for different model selection criteria.}
    \label{fig:grid_precision_at_k}
\end{figure}

Figure \ref{fig:grid_precision_at_k} measures how closely each criterion's ranking matches that defined by the true $\mathrm{MSE}_{\text{test}}$. Again, $\mathrm{MSE}_{\text{train}}$ remains the least reliable. While AIC, AICc, and BIC perform similarly, MDL and Err$_\text{in}$ often show the highest precision. An interesting case is $f_6$, where MDL and Err$_\text{in}$ show similar precision but markedly different average $\mathrm{MSE}_{\text{test}}$ and expression sizes. In this instance, MDL's coding penalty appears overly conservative, potentially down-ranking models that capture the complex shape of $f_6$, whereas the lower parsimony of $\textrm{Err}_{\textrm{in}}$  allows it to retain these more descriptive structures.


\paragraph{Ground-truth hit}

\begin{table}[t]
\centering
\small
\setlength{\tabcolsep}{2pt} 
\begin{tabular}{lrrrrrrrrrrrr}
\toprule
 & $f_1$ & $f_2$ & $f_3$ & $f_4$ & $f_5$ & $f_6$ & $f_7$ & mean & median & std & min $k$ & max $k$ \\
\midrule
$\rm MSE_{train}$ & 77 & 86 & 91 & 77 & 26 & 21 & 54 & 61.71 & 77 & 28.59 & 21 & 91 \\
AIC            & 13 & 11 & 5  & 7  & 7  & 15 & 4  & 8.85  & 7  & 4.18  & 4  & 15 \\
BIC            & 7  & 10 & 5  & 3  & 2  & 11 & 4  & 6.00  & 5  & 3.46  & 2  & 11 \\
AICc           & 11 & 10 & 5  & 6  & 4  & 12 & 4  & 7.42  & 6  & 3.45  & 4  & 12 \\
MDL            & 8  & 2  & 2  & 6  & 4  & 6  & 8  & 5.14  & 6  & 2.54  & 2  & 8 \\
$\textrm{Err}_{\textrm{in}}$        & 13 & 48 & 33 & 20 & 7  & 27 & 74 & 31.71 & 27 & 23.01 & 7  & 74 \\
\bottomrule
\end{tabular}
\caption{Ground-truth hit for different model selection criteria, with row-wise summary statistics.}
\label{tab:gt_hit_firstk_stats}
\end{table}

Table ~\ref{tab:gt_hit_firstk_stats} reports the smallest $k$ for which the generating expression is recovered among the top-$k$ models selected by each criterion. Lower $k$ values indicate superior efficiency in structural recovery. For example, considering the MDL criterion, the ground-truth for benchmark $f_2$ is successfully identified within the first two selected models ($k=2$). 

Aggregated across all benchmarks, MDL and BIC demonstrate the highest recovery efficiency, achieving mean first-hit ranks of 5.14 and 6.00, respectively. In contrast, AIC and AICc exhibit delayed recovery, while $\mathrm{MSE}_{\text{train}}$ performs substantially worse with a mean rank of 61.71. This significant performance gap underscores the detrimental impact of overfitting when model selection is conducted without explicit complexity penalties. 

The in-sample error ($\textrm{Err}_{\textrm{in}}$) criterion yields inconsistent results, particularly for benchmarks $f_2$ and $f_7$, where the ground truth is recovered only at ranks 48 and 74, respectively. This results in a high average rank of 31.71. When cross-referenced with the precision-at-$k$ results in Figure \ref{fig:grid_precision_at_k}, these findings suggest that while $\textrm{Err}_{\textrm{in}}$ can enhance general ranking fidelity, it may down-rank the exact generating expression in favor of near equivalent alternatives that possess competitive bootstrap-corrected errors.

\section{Related work}
\label{sec:related_work}

While SR has first been mentioned as a task solvable by GP, today numerous solution methods for SR exist, including exhaustive~\cite{Bartlett_2022} or enumerative algorithms guided by heuristics\cite{KammererGPTP}, grammatical evolution~\cite{ONeill2004}, sampling from (probabilistic) grammars~\cite{JureBrence}, Monte-carlo tree search and sampling from symbol sequence models with reinforcement learning~\cite{Landajuela_uDSR,sun2023symbolic,xu2024RLSRM}, equation learning via sparsification~\cite{EQL_Sahoo, martius2016extrapolation} or autoencoders, basis function expansion and sparse regression~\cite{FFX,Sindy}, and recently approaches based on embeddings and autoencoders~\cite{tpsr2023,kamienny2023deep}. 
Across these method families, model selection serves as the critical mechanism that filters well-fitting expressions in the search space to a smaller set of solutions. Our empirical analysis is intentionally not algorithm-specific. We have instead opted to analyse the model selection capabilities from a larger set of well-fitting expressions regardless of how this set was produced.

\citet{Ramlan2026} performed a recent study comparing  model selection for GP-based SR based on training error (MSE), AIC, BIC, DL and the scoring metric used by PySR. In contrast to the current work, models were produced using multiple GP runs and 20 real-world datasets from the Penn Machine Learning Benchmarks (PMLB)~\cite{Olson2017PMLB} were used for the empirical analysis instead of synthetic problems. Using these settings produced inconclusive results: no single metric performed reliably across all datasets, whereby metrics that focus mainly on accuracy often led to overfitting, while simplicity-based metrics led to underfitting. This contrasts our results where we see that DL, BIC and the in-sample error estimate, which have stronger penalties, tend to produce smaller models and better test errors. There are multiple causes that might explain the difference in results: likelihood assumptions that are not appropriate for the datasets or the different  process for producing the candidate models among others. We leave this question for future research.



In addition to the criteria studied in this work, Bayesian model selection offers a principled route by comparing models via marginal likelihood (evidence), and the framework of structural risk minimization provides a selection criterion via the VC-dimension. 
A practical challenge with Bayesian model selection is the definition of priors and prior sensitivity. For vague and improper priors over parameters, fractional Bayes factors (FBF) have been proposed for SR \cite{Bartlett_2023,bomarito2022bayesian}. 

\citet{Montana2011} introduced a specific formulation the VC-dimension for structural risk minimization (SRM) for GP-based SR. They compared SRM with AIC and BIC and found empirically better generalization performance for almost all of the tested synthetic benchmark problems. Before that, \citet{Amil2009} mentioned VC dimension for GP-based SR in a theoretical context discussing universal consistency of GP, recently revisited by~\cite{Senn2025}. \citet{Chen2016} observed dramatic gains in generalization performance and smaller models when using SRM. They have only compared to selecting the best-training model, so we do not know how this approach compares to the criteria tested in this work.
\citet{Murari2023} proposed modified robust versions of AIC and BIC and compared with the original forms for GP-based SR. They raise concerns about potential issues arising when using AIC or BIC naively with likelihoods that do not match the datasets, that is using invalid assumptions such as independent or identically distributed errors. This does not affect our results, because we used generated datasets, but is an important limitation of all our tested methods when using model selection criteria with real-world datasets.

\citet{Agapitos2012} used a form of bootstrapping and a weighted fitness criterion combining the loss and a bootstrapped estimate for the variance of the loss value. In contrast to this work, model parameters were not re-fit for each bootstrap sample, ignoring an important component of the variance. 

\section{Conclusion}

We studied model selection for symbolic regression by comparing six criteria, \textit{i.e.} MSE$_{\text{train}}$, AIC, AICc, BIC, MDL, and a parametric bootstrap-corrected in-sample error (Err$_{\text{in}}$) on seven synthetic benchmarks under a controlled evaluation protocol. We assessed selection quality using (i) average MSE$_{\text{test}}$ of the top-$k$ selected models, (ii) precision at $k$ against an MSE$_{\text{test}}$ ranking, (iii) average expression size, and (iv) the first $k$ at which the ground-truth expression is recovered.
Across benchmarks, MDL consistently produced the most compact selections and, in most cases, achieved the lowest (or near-lowest) average MSE$_{\text{test}}$ for small $k$, indicating a favorable simplicity and generalization trade-off. This trend is further reflected in ground-truth recovery, where MDL and BIC recovered the generating expression with the smallest $k$ on average, while $\mathrm{MSE}_{\text{train}}$ was substantially worse, consistent with overfitting when no explicit complexity control is applied. Err$_{\text{in}}$ showed mixed behavior: it can align well with the ranking based on MSE$_{\text{test}}$ in some settings, but exhibited notably late ground-truth recovery on specific benchmarks, suggesting that optimism-correction alone does not guarantee structural recovery within a finite candidate pool.

Taken together, our results support a practical recommendation for GP based SR pipelines that must rank many candidate expressions quickly: when the goal is robust early selection and compact models, MDL (and, secondarily, BIC) is a strong default choice; AIC and AICc may be preferable when slightly larger models are acceptable and the complexity penalty of MDL is overly conservative for a given function family. Because $\rm Err_{in}$ is computationally expensive and did not improve performance in our experiments, we do not recommend it.

A primary limitation of the current study is the reliance on a single noise level used to generate the training data. Future research should investigate how varying noise intensities affect selection performance to assess the robustness of these criteria across different signal-to-noise regimes. Furthermore, integrating these selection criteria directly into the search loop of a symbolic regression algorithm would provide a more comprehensive understanding of their effectiveness in guiding model discovery and preventing overfitting during the evolutionary process.

\begin{acks}
The work of Ali Soltani and Alessandro Lucantonio is supported by the European Union (European Research Council (ERC), ALPS, 101039481). Views and opinions expressed are however those of the author(s) only and do not necessarily reflect those of the European Union or the ERC Executive Agency. Neither the European Union nor the granting authority can be held responsible for them. Computational resources have been partially provided by DeiC National HPC (DeiC-AU-N1-2023030).
F.O.F. is supported by Conselho Nacional de Desenvolvimento Cient\'{i}fico e Tecnol\'{o}gico (CNPq) grant 301596/2022-0.
G.K. acknowledges support by the Austrian Federal Ministry for Climate Action, Environment, Energy, Mobility, Innovation and Technology, the Federal Ministry for Labour and Economy, and the regional government of Upper Austria within the COMET project ProMetHeus (904919) supported by the Austrian Research Promotion Agency (FFG).
ChatGPT was used to check the grammar and improve the style of some parts of the paper.

\end{acks}

\bibliographystyle{ACM-Reference-Format}
\bibliography{sample-base}

@String{Computing = "Computing" }

@String{Computer = "{IEEE} Computer" }

@String{Springer = "Springer-Verlag" }

@BOOK{test,
   author = "Donald E. Knuth",
   title = "Seminumerical Algorithms",
   volume = 2,
   series = "The Art of Computer Programming",
   publisher = "Addison-Wesley",
   address = "Reading, MA",
   edition = "2nd",
   month = "10~" # jan,
   year = "1981",
}

@ArtifactSoftware{R,
    title = {R: A Language and Environment for Statistical Computing},
    author = {{R Core Team}},
    organization = {R Foundation for Statistical Computing},
    address = {Vienna, Austria},
    year = {2019},
    url = {https://www.R-project.org/},
}

@book{Kronberger2024,
  title = {Symbolic Regression},
  DOI = {10.1201/9781315166407},
  publisher = {Chapman and Hall/CRC},
  author = {Kronberger,  Gabriel and Burlacu,  Bogdan and Kommenda,  Michael and Winkler,  Stephan M. and Affenzeller,  Michael},
  year = {2024},
  month = jul 
}

@book{BurnhamAndersonBook2004,
  editor = {Kenneth P. Burnham and David R. Anderson},
  title = {Model Selection and Multimodel Inference},
  ISBN = {9780387953649},
  DOI = {10.1007/b97636},
  publisher = {Springer New York},
  year = {2004}
}

@inproceedings{Bartlett_2023,
author = {Bartlett, Deaglan and Desmond, Harry and Ferreira, Pedro},
title = {Priors for symbolic regression},
year = {2023},
isbn = {9798400701207},
publisher = {Association for Computing Machinery},
address = {New York, NY, USA},
doi = {10.1145/3583133.3596327},
booktitle = {Proceedings of the Companion Conference on Genetic and Evolutionary Computation},
pages = {2402–2411},
numpages = {10},
location = {Lisbon, Portugal},
series = {GECCO '23 Companion}
}

@article{Bartlett_2022,
    author = "Bartlett, Deaglan J. and Desmond, Harry and Ferreira, Pedro G.",
    title = "{Exhaustive Symbolic Regression}",
    journal={IEEE Transactions on Evolutionary Computation}, 
  year={2023},
  volume={},
  number={},
  pages={1-1},
  doi={10.1109/TEVC.2023.3280250},
    eprint = "2211.11461",
    archivePrefix = "arXiv",
    primaryClass = "astro-ph.CO",
}

@Book{Langdon2002,
    author = "William B. Langdon and Riccardo Poli",
    title = "Foundations of Genetic Programming",
    publisher = "Springer-Verlag",
    year = "2002",
    DOI = "10.1007/978-3-662-04726-2",
    size = "274 pages",
}

@book{Koza1992,
	author = {John R. Koza},
	publisher = {MIT Press},
	title = {Genetic Programming: On the Programming of Computers by Means of Natural Selection},
	year = {1992}
}

@article{Friedman1983,
  title = {Multidimensional Additive Spline Approximation},
  volume = {4},
  ISSN = {2168-3417},
  DOI = {10.1137/0904023},
  number = {2},
  journal = {SIAM Journal on Scientific and Statistical Computing},
  publisher = {Society for Industrial & Applied Mathematics (SIAM)},
  author = {Friedman,  Jerome H. and Grosse,  Eric and Stuetzle,  Werner},
  year = {1983},
  month = jun,
  pages = {291–301}
}

@article{Vladislavleva2009,
  title = {Order of Nonlinearity as a Complexity Measure for Models Generated by Symbolic Regression via Pareto Genetic Programming},
  volume = {13},
  ISSN = {1089-778X},
  DOI = {10.1109/tevc.2008.926486},
  number = {2},
  journal = {IEEE Transactions on Evolutionary Computation},
  publisher = {Institute of Electrical and Electronics Engineers (IEEE)},
  author = {Ekaterina J. Vladislavleva and Guido F. Smits  and Dick den Hertog},
  year = {2009},
  month = apr,
  pages = {333–349}
}

@Inbook{Smits2005,
author="Smits, Guido F.
and Kotanchek, Mark",
editor="O'Reilly, Una-May
and Yu, Tina
and Riolo, Rick
and Worzel, Bill",
title="Pareto-Front Exploitation in Symbolic Regression",
bookTitle="Genetic Programming Theory and Practice II",
year="2005",
publisher="Springer US",
address="Boston, MA",
pages="283--299",
doi="10.1007/0-387-23254-0_17",
}

@article{Salustowicz1997,
  title = {Probabilistic Incremental Program Evolution},
  volume = {5},
  ISSN = {1530-9304},
  DOI = {10.1162/evco.1997.5.2.123},
  number = {2},
  journal = {Evolutionary Computation},
  publisher = {MIT Press - Journals},
  author = {Salustowicz,  Rafal and Schmidhuber,  J\"{u}rgen},
  year = {1997},
  month = jun,
  pages = {123–141}
}

@inproceedings{Topchy2001,
author = {Topchy, Alexander and Punch, W. F.},
title = {Faster genetic programming based on local gradient search of numeric leaf values},
year = {2001},
isbn = {1558607749},
publisher = {Morgan Kaufmann Publishers Inc.},
address = {San Francisco, CA, USA},
booktitle = {Proceedings of the 3rd Annual Conference on Genetic and Evolutionary Computation},
pages = {155–162},
numpages = {8},
location = {San Francisco, California},
series = {GECCO'01}
}

@article{arlot2010,
    author = {Sylvain Arlot and Alain Celisse},
    title = {{A survey of cross-validation procedures for model selection}},
    volume = {4},
    journal = {Statistics Surveys},
    number = {none},
    publisher = {Amer. Statist. Assoc., the Bernoulli Soc., the Inst. Math. Statist., and the Statist. Soc. Canada},
    pages = {40 -- 79},
    year = {2010},
    doi = {10.1214/09-SS054},
    URL = {https://doi.org/10.1214/09-SS054}
}

@article{nicolau2021choosing,
  title={Choosing function sets with better generalisation performance for symbolic regression models},
  author={Nicolau, Miguel and Agapitos, Alexandros},
  journal={Genetic programming and evolvable machines},
  volume={22},
  number={1},
  pages={73--100},
  year={2021},
  publisher={Springer}
}

@CONFERENCE{maciel2016measuringcomplexity,
	author = {Maciel, Aron I. and Costa, Ivan G. and Lorena, Ana C.},
	title = {Measuring the complexity of regression problems},
	year = {2016},
	journal = {Proceedings of the International Joint Conference on Neural Networks},
	volume = {2016-October},
	pages = {1450 – 1457},
	doi = {10.1109/IJCNN.2016.7727369},
	url = {https://www.scopus.com/inward/record.uri?eid=2-s2.0-85007146127&doi=10.1109%2fIJCNN.2016.7727369&partnerID=40&md5=89c661dba964f21fe016944b681798e2},
	type = {Conference paper},
	publication_stage = {Final},
	source = {Scopus}
}

@inproceedings{bomarito2022bayesian,
author = {Geoffrey F. Bomarito and Patrik E. Leser and Strauss, N. C. M. and Garbrecht, K. M. and Hochhalter, J. D.},
title = {Bayesian model selection for reducing bloat and overfitting in genetic programming for symbolic regression},
year = {2022},
isbn = {9781450392686},
publisher = {Association for Computing Machinery},
address = {New York, NY, USA},
doi = {10.1145/3520304.3528899},
booktitle = {Proceedings of the Genetic and Evolutionary Computation Conference Companion},
pages = {526–529},
numpages = {4},
location = {Boston, Massachusetts},
series = {GECCO '22}
}

@inbook{ONeill2004,
  title = {Grammatical Evolution by Grammatical Evolution: The Evolution of Grammar and Genetic Code},
  DOI = {10.1007/978-3-540-24650-3_13},
  booktitle = {Genetic Programming},
  publisher = {Springer Berlin Heidelberg},
  author = {O’Neill,  Michael and Ryan,  Conor},
  year = {2004},
  pages = {138–149}
}

@article{Murari2023,
  title = {A practical utility-based but objective approach to model selection for regression in scientific applications},
  volume = {56},
  ISSN = {1573-7462},
  DOI = {10.1007/s10462-023-10591-4},
  number = {S2},
  journal = {Artificial Intelligence Review},
  publisher = {Springer Science and Business Media LLC},
  author = {Murari,  Andrea and Rossi,  Riccardo and Spolladore,  Luca and Lungaroni,  Michele and Gaudio,  Pasquale and Gelfusa,  Michela},
  year = {2023},
  month = oct,
  pages = {2825–2859}
}

@inproceedings{Senn2025,
  series = {GECCO ’25 Companion},
  title = {Model Recovery in Symbolic Regression: Theory,  Conjectures,  and Open Questions},
  DOI = {10.1145/3712255.3734334},
  booktitle = {Proceedings of the Genetic and Evolutionary Computation Conference Companion},
  publisher = {ACM},
  author = {Senn,  Erik-Jan},
  year = {2025},
  month = jul,
  pages = {2556–2562},
  collection = {GECCO ’25 Companion}
}

@article{tpsr2023,
  title={Transformer-based planning for symbolic regression},
  author={Shojaee, Parshin and Meidani, Kazem and Barati Farimani, Amir and Reddy, Chandan},
  journal={Advances in Neural Information Processing Systems},
  volume={36},
  pages={45907--45919},
  year={2023}
}

@inproceedings{kamienny2023deep,
  title={Deep generative symbolic regression with monte-carlo-tree-search},
  author={Kamienny, Pierre-Alexandre and Lample, Guillaume and Lamprier, Sylvain and Virgolin, Marco},
  booktitle={International Conference on Machine Learning},
  pages={15655--15668},
  year={2023},
  organization={PMLR}
}

@inproceedings{
sun2023symbolic,
title={Symbolic Physics Learner: Discovering governing equations via Monte Carlo tree search},
author={Fangzheng Sun and Yang Liu and Jian-Xun Wang and Hao Sun},
booktitle={The Eleventh International Conference on Learning Representations },
year={2023},
url={https://openreview.net/forum?id=ZTK3SefE8_Z}
}

@incollection{KammererGPTP,
  title={Symbolic regression by exhaustive search: Reducing the search space using syntactical constraints and efficient semantic structure deduplication},
  author={Kammerer, Lukas and Kronberger, Gabriel and Burlacu, Bogdan and Winkler, Stephan M and Kommenda, Michael and Affenzeller, Michael},
  booktitle={Genetic programming theory and practice XVII},
  pages={79--99},
  year={2020},
  publisher={Springer}
}

@article{JureBrence,
  title={Probabilistic grammars for equation discovery},
  author={Brence, Jure and Todorovski, Ljup{\v{c}}o and D{\v{z}}eroski, Sa{\v{s}}o},
  journal={Knowledge-Based Systems},
  volume={224},
  pages={107077},
  year={2021},
  publisher={Elsevier}
}

@article{Landajuela_uDSR,
  title={A unified framework for deep symbolic regression},
  author={Landajuela, Mikel and Lee, Chak Shing and Yang, Jiachen and Glatt, Ruben and Santiago, Claudio P and Aravena, Ignacio and Mundhenk, Terrell and Mulcahy, Garrett and Petersen, Brenden K},
  journal={Advances in Neural Information Processing Systems},
  volume={35},
  pages={33985--33998},
  year={2022}
}

@inproceedings{xu2024RLSRM,
  title={Reinforcement symbolic regression machine},
  author={Xu, Yilong and Liu, Yang and Sun, Hao},
  booktitle={The Twelfth International Conference on Learning Representations},
  year={2024}
}

@article{martius2016extrapolation,
  title={Extrapolation and learning equations},
  author={Martius, Georg and Lampert, Christoph H},
  journal={arXiv preprint arXiv:1610.02995},
  year={2016}
}

@incollection{FFX,
  title={FFX: Fast, scalable, deterministic symbolic regression technology},
  author={McConaghy, Trent},
  booktitle={Genetic Programming Theory and Practice IX},
  pages={235--260},
  year={2011},
  publisher={Springer}
}

@article{Sindy,
  title={Discovering governing equations from data by sparse identification of nonlinear dynamical systems},
  author={Brunton, Steven L and Proctor, Joshua L and Kutz, J Nathan},
  journal={Proceedings of the national academy of sciences},
  volume={113},
  number={15},
  pages={3932--3937},
  year={2016},
  publisher={National Academy of Sciences}
}

@inproceedings{Chen2016,
  series = {GECCO ’16},
  title = {Improving Generalisation of Genetic Programming for Symbolic Regression with Structural Risk Minimisation},
  DOI = {10.1145/2908812.2908842},
  booktitle = {Proceedings of the Genetic and Evolutionary Computation Conference 2016},
  publisher = {ACM},
  author = {Chen,  Qi and Xue,  Bing and Shang,  Lin and Zhang,  Mengjie},
  year = {2016},
  month = jul,
  pages = {709–716},
  collection = {GECCO ’16}
}

@inbook{Agapitos2012,
  title = {Controlling Overfitting in Symbolic Regression Based on a Bias/Variance Error Decomposition},
  ISBN = {9783642329371},
  ISSN = {1611-3349},
  DOI = {10.1007/978-3-642-32937-1_44},
  booktitle = {Parallel Problem Solving from Nature - PPSN XII},
  publisher = {Springer Berlin Heidelberg},
  author = {Agapitos,  Alexandros and Brabazon,  Anthony and O’Neill,  Michael},
  year = {2012},
  pages = {438–447}
}

@inbook{Montana2011,
  title = {Penalty Functions for Genetic Programming Algorithms},
  ISBN = {9783642219283},
  ISSN = {1611-3349},
  DOI = {10.1007/978-3-642-21928-3_40},
  booktitle = {Computational Science and Its Applications - ICCSA 2011},
  publisher = {Springer Berlin Heidelberg},
  author = {Montaña,  José L. and Alonso,  César L. and Borges,  Cruz Enrique and de la Dehesa,  Javier},
  year = {2011},
  pages = {550–562}
}

@inbook{Amil2009,
  title = {A Statistical Learning Perspective of Genetic Programming},
  ISBN = {9783642011818},
  ISSN = {1611-3349},
  DOI = {10.1007/978-3-642-01181-8_28},
  booktitle = {Genetic Programming},
  publisher = {Springer Berlin Heidelberg},
  author = {Amil,  Nur Merve and Bredeche,  Nicolas and Gagné,  Christian and Gelly,  Sylvain and Schoenauer,  Marc and Teytaud,  Olivier},
  year = {2009},
  pages = {327–338}
}

@incollection{Ramlan2026,
  author={Fitria Wulandari Ramlan and Gabriel Kronberger and Colm O'Riordan and James McDermott},
  title = {Comparative Analysis of Model Selection Criteria for Symbolic Regression using Genetic Programming},
  pages = {1--18},
  year = {2026},
  publisher = {Springer Nature},
  doi = {10.1007/978-3-032-15635-8_6},
  booktitle = {Computational Intelligence, Proc. of IJCCI 2025, CCIS 2828},
  note = {accepted to be published}

}

@article{Olson2017PMLB,
    author="Olson, Randal S. and La Cava, William and Orzechowski, Patryk and Urbanowicz, Ryan J. and Moore, Jason H.",
    title="PMLB: a large benchmark suite for machine learning evaluation and comparison",
    journal="BioData Mining",
    year="2017",
    month="Dec",
    day="11",
    volume="10",
    number="1",
    pages="36",
    issn="1756-0381",
    doi="10.1186/s13040-017-0154-4",
}

@InProceedings{EQL_Sahoo,
  title = 	 {Learning Equations for Extrapolation and Control},
  author =       {Sahoo, Subham and Lampert, Christoph and Martius, Georg},
  booktitle = 	 {Proceedings of the 35th International Conference on Machine Learning},
  pages = 	 {4442--4450},
  year = 	 {2018},
  editor = 	 {Dy, Jennifer and Krause, Andreas},
  volume = 	 {80},
  series = 	 {Proceedings of Machine Learning Research},
  month = 	 {10--15 Jul},
  publisher =    {PMLR},
  url = 	 {https://proceedings.mlr.press/v80/sahoo18a.html}
}


\end{document}